\documentclass[11pt,a4paper]{article}
\usepackage[hyperref]{emnlp2018}
\usepackage{tipa}
\usepackage{amsmath}
\usepackage{booktabs}
\usepackage{multirow}
\usepackage{times}
\usepackage{latexsym}
\usepackage{graphicx}
\usepackage{float}
\usepackage{subcaption}
\usepackage{placeins}
\usepackage{ragged2e}
\usepackage{paralist}
\usepackage{tikz}
\usepackage{tikz-dependency}
\usepackage[colorinlistoftodos]{todonotes}
\usepackage{url}
\usepackage{cleveref}
\usepackage{lingmacros}
\usepackage{anyfontsize}
\usepackage[flushmargin,hang]{footmisc}

\crefformat{section}{\S#2#1#3} 
\crefformat{subsection}{\S#2#1#3}
\crefformat{subsubsection}{\S#2#1#3}

\aclfinalcopy 

\title{What do character-level models learn about morphology?\\ The case of dependency parsing}

\author{Clara Vania \quad Andreas Grivas\thanks{~~Work done while at the University of Edinburgh.} \quad Adam Lopez \\
  Institute for Language, Cognition and Computation \\
  School of Informatics \\
  University of Edinburgh \\
  {\tt c.vania@ed.ac.uk, andreasgrv@gmail.com, alopez@inf.ed.ac.uk}}

\date{}
\begin{document}
\maketitle

\begin{abstract}
When parsing morphologically-rich languages with neural models, it is beneficial to model input at the character level, and it has been claimed that this is because character-level models learn morphology. We test these claims by comparing character-level models to an oracle with access to explicit morphological analysis on twelve languages with varying morphological typologies. Our results highlight many strengths of character-level models, but also show that they are poor at disambiguating some words, particularly in the face of case syncretism. We then demonstrate that explicitly modeling morphological case improves our best model, showing that character-level models can benefit from targeted forms of explicit morphological modeling.
\end{abstract}

\section{Introduction}

Modeling language input at the character level \cite{ling-EtAl:2015:EMNLP2,kim2015} is effective for many NLP tasks, and often produces better results than modeling at the word level. For parsing, \newcite{ballesteros-dyer-smith:2015:EMNLP} have shown that character-level input modeling is highly effective on morphologically-rich languages, and the three best systems on the 45 languages of the CoNLL 2017 shared task on universal dependency parsing all use character-level models \cite{dozat-qi-manning:2017:K17-3,shi-EtAl:2017:K17-3,bjorkelund-EtAl:2017:K17-3,zeman-EtAl:2017:K17-3}, showing that they are effective across many typologies.

The effectiveness of character-level models in morphologically-rich languages has raised a question and indeed debate about explicit modeling of morphology in NLP. 
\citet{ling-EtAl:2015:EMNLP2} propose that ``prior information regarding morphology ... among others, should be incorporated'' into character-level models, while \citet{chung-cho-bengio:2016:P16-1} counter that it is ``unnecessary to consider these prior information'' when modeling characters. Whether we need to explicitly model morphology is a question whose answer has a real cost: as \newcite{ballesteros-dyer-smith:2015:EMNLP} note, morphological annotation is expensive, and this expense could be reinvested elsewhere if the predictive aspects of morphology are learnable from strings.

Do character-level models learn morphology? We view this as an empirical claim requiring empirical evidence. The claim has been tested implicitly by comparing character-level models to word lookup models \cite{qian-qiu-huang:2016:P16-11,belinkov-EtAl:2017:Long}. In this paper, we test it explicitly, asking how character-level models compare with an oracle model with access to morphological annotations. This extends experiments showing that character-aware language models in Czech and Russian benefit substantially from oracle morphology \citep{P17-1184}, but here we focus on dependency parsing  (\textsection\ref{sec:parser})---a task that benefits substantially from morphological knowledge---and we experiment with twelve languages using a variety of techniques to probe our models. 

Our summary finding is that character-level models lag the oracle in nearly all languages (\textsection\ref{sec:experiments}). The difference is small, but suggests that there is value in modeling morphology. When we tease apart the results by part of speech and dependency type, we trace the difference back to the character-level model's inability to disambiguate words even when encoded with arbitrary context (\textsection\ref{sec:analysis}). Specifically, it struggles with \emph{case syncretism}, in which noun case---and thus syntactic function---is ambiguous. We show that the oracle relies on morphological case, and that a character-level model provided \emph{only} with morphological case rivals the oracle, even when case is provided by another predictive model (\textsection\ref{sec:case-matters}). Finally, we show that the crucial morphological features vary by language (\textsection\ref{sec:understanding-head-selection}).

\section{Dependency parsing model}
\label{sec:parser}

We use a neural graph-based dependency parser combining elements of two recent models \citep{kiperwasser2016,zhang-cheng-lapata:2017:EACLlong}. Let $w = w_1, \dots, w_{|w|}$ be an input sentence of length $|w|$ and let $w_0$ denote an artificial \textsc{Root} token. We represent the $i$th input token $w_i$ by concatenating its \textit{word representation} (\textsection\ref{sec:word-reps}), $\textbf{e}(w_i)$ and part-of-speech (POS) representation, $\textbf{p}_i$.\footnote{This combination yields the best labeled accuracy according to \newcite{ballesteros-dyer-smith:2015:EMNLP}.} Using a semicolon $(;)$ to denote vector concatenation, we have:
\begin{align}
\label{eq:in-rep}
\textbf{x}_i = [\textbf{e}(w_i);\textbf{p}_i]
\end{align}
We call $\textbf{x}_i$ the \textit{embedding} of $w_i$ since it depends on context-independent word and POS representations. We obtain a context-sensitive \textit{encoding} $\textbf{h}_i$ with a bidirectional LSTM (bi-LSTM), which concatenates the hidden states of a forward and backward LSTM at position $i$. Using $\textbf{h}_i^f$ and $\textbf{h}_i^b$ respectively to denote these hidden states, we have:
\begin{align}
\label{eq:enc-rep}
\textbf{h}_i = [\textbf{h}_i^f;\textbf{h}_i^b]
\end{align}
We use $\textbf{h}_i$ as the final input representation of $w_i$.

\subsection{Head prediction}

For each word $w_i$, we compute a distribution over all other word positions $j \in \{0,...,|w|\}/i$ denoting the probability that $w_j$ is the headword of $w_i$.
\begin{align}
\label{eq:head-pred}
P_{head}(w_j \mid w_i,w) = \frac{\exp(a(\textbf{h}_i, \textbf{h}_j))}{\sum_{j'=0}^{|w|} \exp(a(\textbf{h}_i, \textbf{h}_{j'}))}
\end{align}
Here, $a$ is a neural network that computes an association between $w_i$ and $w_j$ using model parameters $\textbf{U}_a, \textbf{W}_a,$ and $\textbf{v}_a$.
\begin{align}
\label{eq:ass-score}
a(\textbf{h}_i, \textbf{h}_j) = \textbf{v}_a \tanh(\textbf{U}_a \textbf{h}_i + \textbf{W}_a \textbf{h}_j)
\end{align}

\subsection{Label prediction}

Given a head prediction for word $w_i$, we predict its syntactic label $\ell_k \in L$ using a similar network. 
\begin{align}
\label{eq:label-pred}
P_{label}(\ell_k \mid w_i, w_j, w) = \frac{\exp(f(\textbf{h}_i, \textbf{h}_j)[k])}{\sum_{k'=1}^{|L|} \exp(f(\textbf{h}_i, \textbf{h}_{j})[k'])}
\end{align}
where $L$ is the set of output labels and $f$ is a function that computes label score using model parameters $\textbf{U}_\ell, \textbf{W}_\ell,$ and $\textbf{V}_\ell$:
\begin{align}
\label{eq:lbl-score}
f(\textbf{h}_i, \textbf{h}_j) = \textbf{V}_\ell \tanh(\textbf{U}_\ell \textbf{h}_i + \textbf{W}_\ell \textbf{h}_j)
\end{align}
The model is trained to minimize the summed cross-entropy losses of both head and label prediction. At test time, we use the Chu-Liu-Edmonds \cite{chu-liu-1965,edmonds} algorithm to ensure well-formed, possibly non-projective trees.

\subsection{Computing word representations}
\label{sec:word-reps}

We consider several ways to compute the word representation $\textbf{e}({w_i})$ in Eq. \ref{eq:in-rep}:

\begin{asparaitem}
\item[\textbf{word}.] Every word type has its own learned vector representation.
\item[\textbf{char-lstm}.] Characters are composed using a bi-LSTM \citep{ling-EtAl:2015:EMNLP2}, and the final states of the forward and backward LSTMs are concatenated to yield the word representation. 
\item[\textbf{char-cnn}.] Characters are composed using a convolutional neural network \citep{kim2015}.
\item[\textbf{trigram-lstm}.] Character trigrams are composed using a bi-LSTM, an approach that we previously found to be effective across typologies \citep{P17-1184}.
\item[\textbf{oracle}.] We treat the morphemes of a morphological annotation as a sequence and compose them using a bi-LSTM. We only use universal inflectional features defined in the UD annotation guidelines. For example, the morphological annotation of ``chases'' is \texttt{$\langle$chase, person=3rd, num-SG, tense=Pres$\rangle$}.
\end{asparaitem}

For the remainder of the paper, we use the name of model as shorthand for the dependency parser that uses that model as input (Eq. \ref{eq:in-rep}).

\section{Experiments}\label{sec:experiments}

\paragraph{Data} We experiment on twelve languages with varying morphological typologies (Table \ref{tab:train-stat}) in the Universal Dependencies (UD) treebanks version 2.0 \cite{UD2.0}.\footnote{For Russian we use the UD\_Russian\_SynTagRus treebank, and for all other languages we use the default treebank.} Note that while Arabic and Hebrew follow a root \& pattern typology, their datasets are unvocalized, which might reduce the observed effects of this typology. Following common practice, we remove language-specific dependency relations and multiword token annotations. We use gold sentence segmentation, tokenization, universal POS (UPOS), and morphological (XFEATS) annotations provided in UD.

\paragraph{Implementation and training} Our Chainer  \cite{chainer_learningsys2015} implementation encodes words (Eq.~\ref{eq:enc-rep}) in two-layer bi-LSTMs with 200 hidden units, and uses 100 hidden units for head and label predictions (output of Eqs. 4 and 6). We set batch size to 16 for char-cnn and 32 for other models following a grid search. We apply dropout to the embeddings (Eq.~\ref{eq:in-rep}) and the input of the head prediction. We use Adam optimizer with initial learning rate 0.001 and clip gradients to 5, and train all models for 50 epochs with early stopping. For the word model, we limit our vocabulary to the 20K most frequent words, replacing less frequent words with an unknown word token. The char-lstm, trigram-lstm, and oracle models use a one-layer bi-LSTM with 200 hidden units to compose subwords. For char-cnn, we use the small model setup of \newcite{kim2015}. 

\begin{table}[t]
\centering
\begin{tabular}{l r r r}
\toprule
Languages & \#sents & \#tokens & type/token \\
 & (K) & (K) & ratio (\%) \\
\midrule
Finnish & 12.2 & 162.6 & 28.5 \\
Turkish & 3.7 & 38.1 & 33.6 \\
\midrule
Czech & 68.5 & 1173.3 & 9.5 \\
English & 12.5 & 204.6 & 8.1 \\
German & 14.1 & 269.6 & 17.7 \\
Hindi & 13.3 & 281.1 & 6 \\
Portuguese & 8.3 & 206.7 & 11.7 \\
Russian & 48.8 & 870 & 11.4 \\
Spanish & 14.2 & 382.4 & 11.1 \\
Urdu & 4.0 & 108.7 & 8.8 \\
\midrule
Arabic & 6.1 & 223.9 & 10.3 \\
Hebrew & 5.2 & 137.7 & 11.7 \\
\bottomrule
\end{tabular}
\caption{Training data statistics. Languages are grouped by their dominant morphological processes, from top to bottom: agglutinative, fusional, and root \& pattern.}
\label{tab:train-stat}
\end{table}

\begin{table*}[ht]
\centering
\begin{tabular}{lllllllll|ll|l}
\toprule
Model $\rightarrow$ & \multicolumn{2}{c}{word} & \multicolumn{2}{c}{char-lstm} & \multicolumn{2}{c}{char-cnn} & \multicolumn{2}{c}{trigram-lstm} & \multicolumn{2}{|c|}{oracle} & o/c \\
 \cmidrule{2-12}
$\downarrow$ Language & UAS & LAS & UAS & LAS & UAS & LAS & UAS & LAS & UAS & LAS & LAS \\
  \midrule
  Finnish & 85.7 & 80.8 & \textbf{90.6} & \textbf{88.4} & 89.9 & 87.5 & 89.7 & 87.0 & 
  	90.6 & 88.8 & +0.4 \\
  Turkish & 71.4 & 61.6 & \textbf{74.7} & \textbf{68.6} & 74.4 & 67.9 & 73.2 & 65.9 & 
  	75.3 & 69.5 & +0.9 \\
  \midrule
  Czech & 92.6 & 89.3 & \textbf{93.5} & \textbf{90.6} & \textbf{93.5} & \textbf{90.6} & 
  	92.7 & 89.2 & 94.3 & 92.0 & +1.4 \\
  English & 90.6 & 88.9 & 91.3 & 89.4 & \textbf{91.7} & \textbf{90.0} & 90.4 & 88.5 & 
  	91.7 & 89.9 & +0.5 \\
  German & \textbf{88.1} & \textbf{84.5} & 88.0 & \textbf{84.5} & 87.8 & 84.4 & 87.1 & 
  	83.5 & 88.8 & 86.5 & +2.0 \\
  Hindi & \textbf{95.8} & 93.1 & 95.7 & \textbf{93.3} & 95.7 & 93.2 & 93.4 & 89.8 & 
  	95.9 & 93.3 & - \\
  Portuguese & 87.4 & 85.5 & \textbf{87.8} & \textbf{86.0} & 87.7 & \textbf{86.0} & 
  	86.7 & 84.8 & 88.0 & 86.5 & +0.5 \\
  Russian & 92.4 & 90.1 & \textbf{94.0} & \textbf{92.4} & 93.8 & 92.1 & 92.0 & 89.5 & 
  	94.4 & 93.3 & +0.9 \\
  Spanish & 89.4 & 86.9 & 89.8 & \textbf{87.4} & \textbf{90.0} & 87.3 & 88.6 & 85.5 & 
  	90.0 & 87.7 & +0.3 \\
  Urdu & 91.1 & 87.0 & 91.2 & 87.1 & \textbf{91.3} & \textbf{87.2} & 88.6 & 83.5 & 90.9 
  	& 87.0 & -0.1 \\
  \midrule
  Arabic & 75.5 & 70.9 & \textbf{76.7} & 72.1 & 76.6 & \textbf{72.2} & 74.6 & 68.9 & 76.7 & 	72.7 & +0.6 \\
  Hebrew & \textbf{73.5} & \textbf{69.8} & 73.4 & \textbf{69.8} & 73.3 & \textbf{69.8} & 71.3 & 67.1 	& 73.3 & 70.0 & +0.2\\
\bottomrule
\end{tabular}
\caption{Unlabeled Attachment Score (UAS) and Labeled Attachment Score (LAS) on \textbf{test set}. The best accuracy for each language is highlighted in \textbf{bold} for all models, and for all non-oracle models. \textbf{o/c:} LAS improvement from char-lstm to oracle.}
\label{tab:test-res}
\end{table*}

\paragraph{Parsing Results}

Table \ref{tab:test-res} presents test results for every model on every language, establishing three results. First, they support previous findings that character-level models outperform word-based models---indeed, the char-lstm model outperforms the word model on LAS for all languages except Hindi and Urdu for which the results are identical.\footnote{Note that Hindi and Urdu are mutually intelligible.} Second, they establish strong baselines for the character-level models: the char-lstm generally obtains the best parsing accuracy, closely followed by char-cnn. Third, they demonstrate that character-level models rarely match the accuracy of an oracle model with access to explicit morphology. This reinforces a finding of \citet{P17-1184}: character-level models are effective tools, but they do not learn everything about morphology, and they seem to be closer to oracle accuracy in agglutinative rather than in fusional languages.

\begin{table}[t]
\centering
\begin{tabular}{l c c c}
\toprule
\multirow{2}{*}{Language} & dev & \multicolumn{2}{c}{LAS improvement} \\
\cmidrule{3-4}
 & \%OOV & non-OOV & OOV \\
\midrule
Finnish & 23.0 & 6.8 & 17.5 \\
Turkish & 24.0 & 4.6 & 13.5 \\
\midrule
Czech & 5.8 & 1.4 & 3.9 \\
English & 6.8 & 0.7 & 5.2 \\
German & 9.7 & 0.9 & 0.7 \\
Hindi & 4.3 & 0.2 & 0.0 \\
Portuguese & 8.1 & 0.3 & 1.3 \\
Russian & 8.4 & 2.1 & 6.9 \\
Spanish & 7.0 & 0.4 & 0.7 \\
\midrule
Arabic & 8.0 & 1.2 & 7.3 \\
Hebrew & 9.0 & 0.2 & 1.3 \\
\bottomrule
\end{tabular}
\caption{LAS improvements (char-lstm $-$ word) for non-OOV and OOV words on development set.}
\label{tab:oov-res}
\end{table}

\section{Analysis}\label{sec:analysis}

\subsection{Why do characters beat words?}
In character-level models, orthographically similar words share many parameters, so we would expect these models to produce good representations of OOV words that are morphological variants of training words. Does this effect explain why they are better than word-level models? 

\paragraph{Sharing parameters helps with both seen and unseen words}
Table~\ref{tab:oov-res} shows how the character model improves over the word model for both non-OOV and OOV words. On the agglutinative languages Finnish and Turkish, where the OOV rates are 23\% and 24\% respectively, we see the highest LAS improvements, and we see especially large improvements in accuracy of OOV words. However, the effects are more mixed in other languages, even with relatively high OOV rates. In particular, languages with rich morphology like Czech, Russian, and (unvocalised) Arabic see more improvement than languages with moderately rich morphology and high OOV rates like Portuguese or Spanish. This pattern suggests that parameter sharing between pairs of \emph{observed} training words can also improve parsing performance. For example, if ``dog'' and ``dogs'' are observed in the training data, they will share activations in their context \textit{and} on their common prefix.

\begin{table*}[!ht]
\centering
  \begin{tabular}{llccccc|c}
  \toprule
  Language & Model & ADJ & NOUN & PRON & PROPN & VERB & Overall \\
  \midrule
  Finnish & \%tokens & 8.1 & 32.5 & 8.2 & 6.7 & 16.1 & - \\
   \cmidrule{2-8}
   & char-lstm & 89.2 & 82.1 & 88.1 & 84.5 & 88.4 & 87.7 \\
   & oracle & 90.3 & 83.3 & 89.5 & 86.2 & 89.3 & 88.5 \\
   & diff & +1.1 & +1.2 & +1.4 & +1.7 & +0.9 & +0.8 \\
   \midrule
  Czech & \%tokens & 14.9 & 28.7 & 3.6 & 6.3 & 10.7 & - \\
   \cmidrule{2-8}
   & char-lstm & 94.2 & 83.6 & 85.3 & 84.3 & 90.7 & 91.2 \\
   & oracle & 94.8 & 87.5 & 88.5 & 86.8 & 91.1 & 92.5 \\
   & diff & +0.6 & +3.9 & +3.2 & +2.5 & +0.4 & +1.3 \\
   \midrule
  German & \%tokens & 7.6 & 20.4 & 9.5 & 5.6 & 12.1 & - \\
   \cmidrule{2-8}
   & char-lstm & 88.4 & 81.4 & 86.0 & 82.4 & 85.2 & 87.5 \\
   & oracle & 89.1 & 87.1 & 93.2 & 84.4 & 86.3 & 89.7 \\
   & diff & +0.7 & +5.7 & +7.2 & +2.0 & +1.1 & +2.2 \\
   \midrule
  Russian & \%tokens & 12.2 & 29.3 & 6.1 & 4.6 & 13.7 & - \\
   \cmidrule{2-8}
   & char-lstm & 93.2 & 86.7 & 92.0 & 80.2 & 88.5 & 91.6 \\
   & oracle & 93.7 & 88.8 & 93.3 & 86.4 & 88.9 & 92.6 \\
   & diff & +0.5 & +2.1 & +1.3 & +6.2 & +0.4 & +1.0 \\
   \bottomrule
  \end{tabular}
\caption{Labeled accuracy for different parts of speech on development set.}
\label{tab:pos-res}
\end{table*}

\subsection{Why do morphemes beat characters?}

Let's turn to our main question: what do character-level models learn about morphology? To answer it, we compare the oracle model to char-lstm, our best character-level model.

\paragraph{Morphological analysis disambiguates words} In the oracle,  morphological annotations disambiguate some words that the char-lstm must disambiguate from context. Consider these Russian sentences from \newcite{baerman-brown-corbett-2005}:
\enumsentence{
\vskip-.5em
\label{ex-amb-1}
   Ma\v{s}a \v{c}itaet \underline{pis\textceltpal mo} \\
   Masha reads \underline{letter} \\
   `\textit{Masha reads a \underline{letter}.}'
}
\enumsentence{
\vskip-.5em
\label{ex-amb-2}
	Na stole le\v{z}it \underline{pis\textceltpal mo} \\
    on table lies \underline{letter} \\
    `\textit{There's a \underline{letter} on the table.}'
}
\textit{Pis\textceltpal mo} (``letter'') acts as the subject in (\ref{ex-amb-1}), and as object in (\ref{ex-amb-2}). This knowledge is available to the oracle via morphological case: in (\ref{ex-amb-1}), the case of \textit{pis\textceltpal mo} is nominative and in (\ref{ex-amb-2}) it is accusative. Could this explain why the oracle outperforms the character model?

\begin{figure}[t]
\includegraphics[width=.95\linewidth]{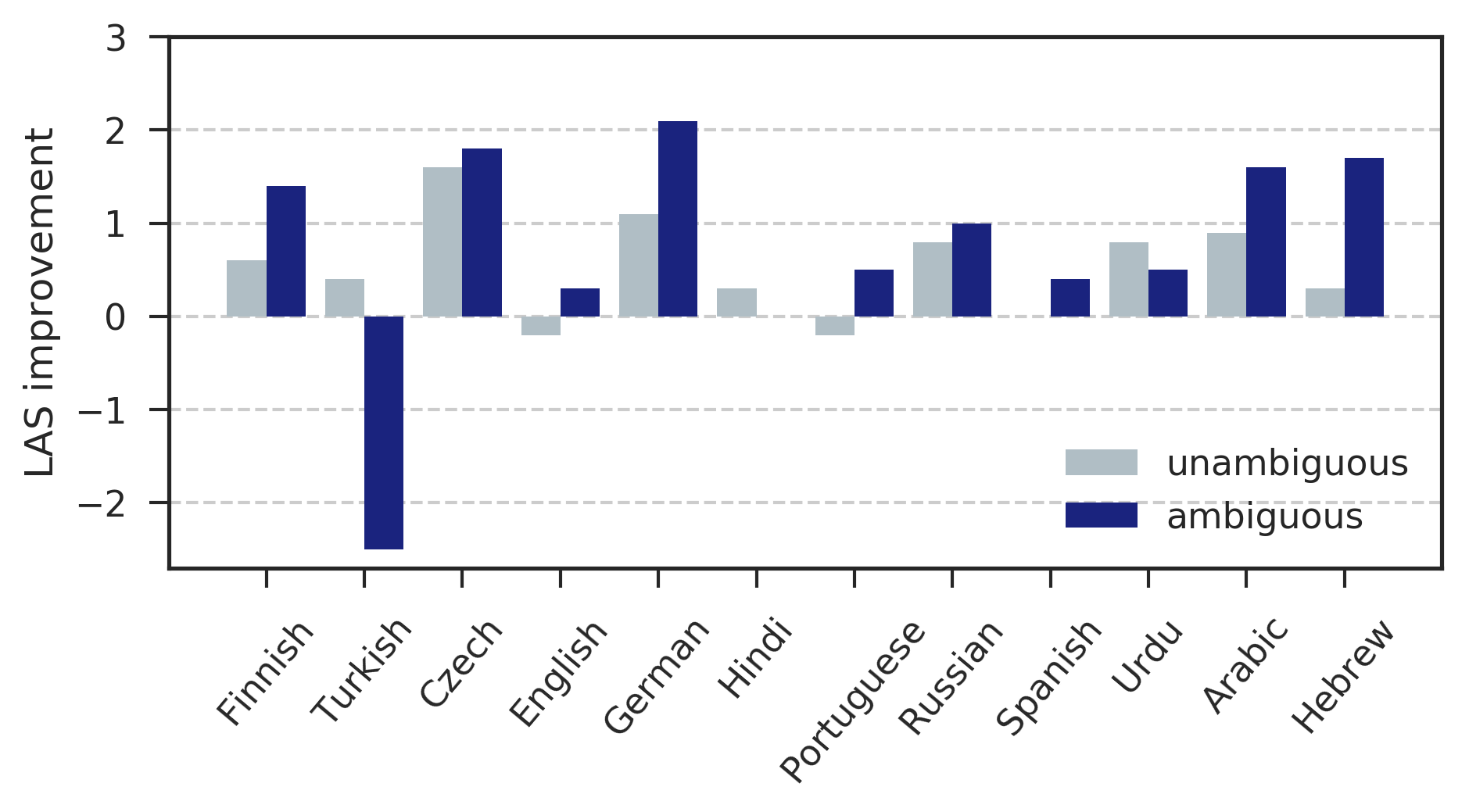}
\caption{LAS improvements (oracle $-$ char-lstm) for ambiguous and unambiguous words on development set.}
\label{fig:amb}
\end{figure}

To test this, we look at accuracy for word types that are \emph{empirically} ambiguous---those that have more than one morphological analysis in the training data. Note that by this definition, some ambiguous words will be seen as unambiguous, since they were seen with only one analysis. To make the comparison as fair as possible, we consider only words that were observed in the training data. Figure \ref{fig:amb} compares the improvement of the oracle on ambiguous and seen unambiguous words, and as expected we find that handling of ambiguous words improves with the oracle in almost all languages. The only exception is Turkish, which has the least training data.

\vskip-1em

\paragraph{Morphology helps for nouns}
Now we turn to a more fine-grained analysis conditioned on the annotated part-of-speech (POS) of the \textit{dependent}. We focus on four languages where the oracle strongly outperforms the best character-level model on the development set: Finnish, Czech, German, and Russian.\footnote{This is slightly different than on the test set, where the effect was stronger in Turkish than in Finnish. In general, we found it difficult to draw conclusions from Turkish, possibly due to the small size of the data.} We consider five POS categories that are frequent in all languages and consistently annotated for morphology in our data: adjective (ADJ), noun (NOUN), pronoun (PRON), proper noun (PROPN), and verb (VERB). 

Table \ref{tab:pos-res} shows that the three noun categories---ADJ, PRON, and PROPN---benefit substantially from oracle morphology, especially for the three fusional languages: Czech, German, and Russian. 

\paragraph{Morphology helps for subjects and objects}
We analyze results by the dependency type of the \textit{dependent}, focusing on types that interact with morphology: \textit{root}, nominal subjects (\textit{nsubj}), objects (\textit{obj}), indirect objects (\textit{iobj}), nominal modifiers (\textit{nmod}), adjectival modifier (\textit{amod}), obliques (\textit{obl}), and (syntactic) case markings (\textit{case}). 

\begin{figure*}[t]
	\includegraphics[width=\linewidth]{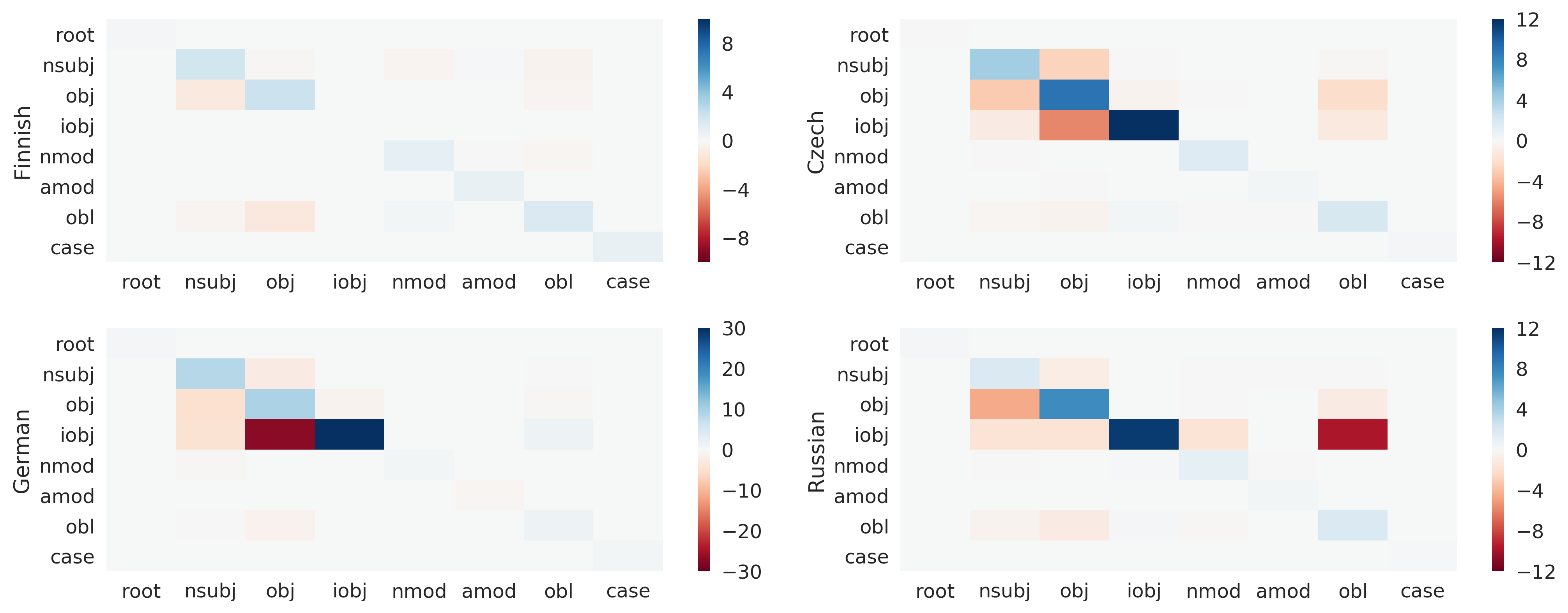}
    \caption{Heatmaps of the difference between oracle vs. char-lstm confusion matrices for label prediction when both head predictions are correct (\textbf{x-axis}: predicted labels; \textbf{y-axis}: gold labels). Blue cells have higher oracle values, red cells have higher char-lstm values.}
	\label{fig:deplab-res}
\end{figure*}

\begin{figure}[t]
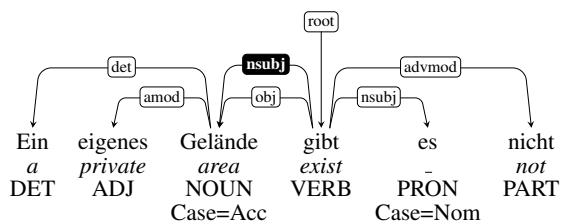

\scalebox{.8}{\input
\begin{dependency}
\begin{deptext}[column sep=0.2cm]
  Ein \& eigenes \& Gel\"ande \& gibt \& es \& nicht \\
  \textit{a} \& \textit{private} \& \textit{area} \& \textit{exist} \& \_ \& \textit{not} \\
  DET \& ADJ \& NOUN \& VERB \& PRON \& PART \\
   \& \& Case=Acc \& \& Case=Nom \& \\
  \end{deptext}
  \depedge{3}{1}{det}
  \depedge{3}{2}{amod}
  \depedge{4}{3}{obj}
  \depedge[label theme=night,edge unit distance=6ex]{4}{3}{nsubj}
  \depedge{4}{6}{advmod}
  \depedge{4}{5}{nsubj}
  \deproot{4}{root}
\end{dependency}}
\caption{A sentence which the oracle parses perfectly (shown in white) and the char-lstm predicts an incorrect label (shown in black).}
\label{fig:lab-ex}
\end{figure}

Figure~\ref{fig:deplab-res} shows the differences in the confusion matrices of the char-lstm and oracle for those words on which both models correctly predict the head. The differences on Finnish are small, which we expect from the similar overall LAS of both models. But for the fusional languages, a pattern emerges: the char-lstm consistently underperforms the oracle on nominal subject, object, and indirect object dependencies---labels closely associated with noun categories. From inspection, it appears to frequently mislabel objects as nominal subjects when the dependent noun is morphologically ambiguous. For example, in the sentence of Figure~\ref{fig:lab-ex}, \textit{Gel\"ande} (``terrain'') is an object, but the char-lstm incorrectly predicts that it is a nominal subject. In the training data, \textit{Gel\"ande} is ambiguous: it can be accusative, nominative, or dative. 

In German, the char-lstm frequently confuses objects and indirect objects. By inspection, we found 21 mislabeled cases, where 20 of them would likely be correct if the model had access to morphological case (usually dative). In Czech and Russian, the results are more varied: indirect objects are frequently mislabeled as objects, obliques, nominal modifiers, and nominal subjects. We note that indirect objects are relatively rare in these data, which may partly explain their frequent mislabeling.

\section{Characters and case syncretism}\label{sec:case-matters}

So far, we've seen that for our three fusional languages---German, Czech, and Russian---the oracle strongly outperforms a character model on nouns with ambiguous morphological analyses, particularly on core dependencies: nominal subjects, objects and indirect objects. Since the nominative, accusative, and dative morphological cases are strongly (though not perfectly) correlated with these dependencies, it is easy to see why the morphologically-aware oracle is able to predict them so well. We hypothesized that these cases are more challenging for the character model because these languages feature a high degree of \textit{syncretism}---functionally distinct words that have the same form---and in particular case syncretism. For example, referring back to examples (\ref{ex-amb-1}) and (\ref{ex-amb-2}), the character model must disambiguate \textit{pis\textceltpal mo} from its context, whereas the oracle can directly disambiguate it from a feature of the word itself.\footnote{We are far from first to observe that morphological case is important to parsing: \newcite{seeker2013} observe the same for non-neural parsers.}

To understand this, we first designed an experiment to see whether the char-lstm could successfully disambiguate \emph{noun} case, using a method similar to \cite{belinkov-EtAl:2017:Long}. We train a neural classifier that takes as input a word representation from the trained parser and predicts a  morphological feature of that word---for example that its case is nominative (\texttt{Case=Nom}). The classifier is a feedforward neural network with one hidden layer, followed by a ReLU non-linearity. We consider two representations of each word: its embedding ($\textbf{x}_i$; Eq. \ref{eq:in-rep}) and its encoding ($\textbf{h}_i$; Eq. \ref{eq:enc-rep}). To understand the importance of case, we consider it alongside number and gender features as well as whole feature bundles.

\paragraph{The oracle relies on case}
Table \ref{tab:classifier-res} shows the results of morphological feature classification on Czech; we found very similar results in German and Russian (Appendix \ref{sec:sup-tagging}). The oracle embeddings have almost perfect accuracy---and this is just what we expect, since the representation only needs to preserve information from its input. The char-lstm embeddings perform well on number and gender, but less well on case. This results suggest that the character-level models still struggle to learn case when given only the input text. Comparing the char-lstm with a baseline model which predicts the most frequent feature for each type in the training data, we observe that both of them show similar trends even though character models slightly outperforms the baseline model.

\begin{table}[!t]
\centering
\setlength\tabcolsep{5pt}
  \begin{tabular}{l c c c c c}
  \toprule
  \multirow{2}{*}{Feature} & \multirow{2}{*}{baseline} & \multicolumn{2}{c}{embedding} & \multicolumn{2}{c}{encoder} 	\\
   \cmidrule{3-6}
   & & char & oracle & char & oracle \\
   \midrule
    Case  & 71.1 & 74.4 & \textbf{100} & 86.5 & \textbf{98.6} \\
    Gender & 92.9 & 98.1 & \textbf{100} & \textbf{71.2} & 58.6 \\
    Number & 88.9 & 94.7 & \textbf{100} & 84.2 & \textbf{84.8} \\
    All & 70.4 & 72.5 & \textbf{99.9} & \textbf{58.1} & 50.2 \\
   \bottomrule
  \end{tabular}
\caption{Morphological tagging accuracy from representations using the char-lstm and oracle embedding and encoder representations in Czech. Baseline simply chooses the most frequent tag. \emph{All} means we concatenate all annotated features in UD as one tag.}
\label{tab:classifier-res}
\end{table}

The classification results from the encoding are particularly interesting: the oracle still performs very well on morphological case, but less well on other features, even though they appear in the input. In the character model, the accuracy in morphological prediction also degrades in the encoding---\emph{except} for case, where accuracy on case improves by 12\%. 

These results make intuitive sense: representations learn to preserve information from their input that is useful for subsequent predictions. In our parsing model, morphological case is very useful for predicting dependency labels, and since it is present in the oracle's input, it is passed almost completely intact through each representation layer. The character model, which must disambiguate case from context, draws as much additional information as it can from surrounding words through the LSTM encoder. But other features, and particularly whole feature bundles, are presumably less useful for parsing, so neither model preserves them with the same fidelity.\footnote{This finding is consistent with \newcite{ballesteros:2013:SPMRL} which performed careful feature analysis on morphologically rich languages and found that lemma and case features provide the highest improvement in a non-neural transition based parser compared to other features.}

\paragraph{Explicitly modeling case improves parsing accuracy}

Our analysis indicates that case is important for parsing, so it is natural to ask: Can we improve the neural model by explicitly modeling case? To answer this question, we ran a set of experiments, considering two ways to augment the char-lstm with case information: multitask learning \citep[MTL;][]{Caruana1997} and a pipeline model in which we augment the char-lstm model with either predicted or gold case. For example, we use \texttt{$\langle$p, i, z, z, a, Nom$\rangle$} to represent \textit{pizza} with nominative case. For MTL, we follow the setup of \citet{sogaard-acl16} and \citet{coavoux-eacl17}. We increase the biLSTMs layers from two to four and use the first two layers to predict morphological case, leaving out the other two layers specific only for parser. For the pipeline model, we train a morphological tagger to predict morphological case (Appendix \ref{sec:morph-tagger}). This tagger does not share parameters with the parser.
\begin{table}[t]
\centering
  \begin{tabular}{l l c c c}
  \toprule
  Language & Input & Dev & Test \\
  \midrule
  Czech & char & 91.2 & 90.6 \\
        & char (multi-task) & 91.6 & 91.0 \\ 
        & char + predicted case & \textbf{92.2} & \textbf{91.8} \\
        \cmidrule{2-4}
        & char + gold case & 92.3 & 91.9 \\
        & oracle & 92.5 & 92.0 \\
  \midrule
  German & char & 87.5 & 84.5 \\
         & char (multi-task) & \textbf{87.9} & 84.4 \\
         & char + predicted case & 87.8 & \textbf{86.4} \\
         \cmidrule{2-4}
         & char + gold case & 90.2 & 86.9 \\
         & oracle & 89.7 & 86.5 \\
  \midrule
  Russian & char & 91.6 & 92.4 \\
          & char (multi-task) & 92.2 & 92.6 \\
          & char + predicted case & \textbf{92.5} & \textbf{93.3} \\
          \cmidrule{2-4}
          & char + gold case & 92.8 & 93.5 \\
          & oracle & 92.6 & 93.3 \\
  \bottomrule
  \end{tabular}
  \caption{LAS results when case information is added. We use \textbf{bold} to highlight the best results for models without explicit access to gold annotations.}
  \label{tab:adding-case-exp}
\end{table}

Table \ref{tab:adding-case-exp} summarizes the results on Czech, German, and Russian. We find augmenting the char-lstm model with either oracle or predicted case improve its accuracy, although the effect is different across languages. The improvements from predicted case results are interesting, since in non-neural parsers, predicted case usually harms accuracy \citep{Tsarfaty:2010:SPM:1868771.1868772}. However, we note that our taggers use gold POS, which might help. The MTL models achieve similar or slightly better performance than the character-only models, suggesting that supplying case in this way is beneficial. Curiously, the MTL parser is worse than the the pipeline parser, but the MTL case tagger is better than the pipeline case tagger (Table \ref{tab:tag-acc}). This indicates that the MTL model must learn to encode case in the model's representation, but must not learn to effectively use it for parsing. Finally, we observe that augmenting the char-lstm with either gold or predicted case improves the parsing performance for all languages, and indeed closes the performance gap with the full oracle, which has access to \emph{all} morphological features. This is especially interesting, because it shows using carefully targeted linguistic analyses can improve accuracy as much as wholesale linguistic analysis.

\begin{table}
\centering
	\begin{tabular}{l c c c c c}
    \toprule
	\multirow{2}{*}{Language} & \multirow{2}{*}{\%case} & \multicolumn{2}{c}{Dev} & \multicolumn{2}{c}{Test} \\
    \cmidrule{3-6}
     & & PL & MT & PL & MT \\
    \midrule
    Czech & 66.5 & 95.4 & \textbf{96.7} & 95.2 & \textbf{96.6} \\
    German & 36.2 & \textbf{92.6} & 92.0 & 90.8 & \textbf{91.4} \\
  	Russian & 55.8 & 95.8 & \textbf{96.5} & 95.9 & \textbf{96.5} \\
    \bottomrule
	\end{tabular}
    \caption{Case accuracy for case-annotated tokens, for pipeline (\textbf{PL}) vs. multitask (\textbf{MT}) setup. \textbf{\%case} shows percentage of training tokens annotated with case.}
    \label{tab:tag-acc}
\end{table}

\section{Understanding head selection}\label{sec:understanding-head-selection}
\label{sec:attn}

The previous experiments condition their analysis on the \textit{dependent}, but dependency is a relationship between dependents and heads. We also want to understand the importance of morphological features to the \textit{head}. Which morphological features of the head are important to the oracle? 

\paragraph{Composing features in the oracle}
To see which morphological features the oracle depends on when making predictions, we augmented our model with a \textbf{gated  attention mechanism} following \newcite{kuncoro-EtAl:2017:EACLlong}. Our new model attends to the morphological features of candidate head $w_j$ when computing its association with dependent $w_i$ (Eq.~\ref{eq:head-pred}), and morpheme representations are then scaled by their \textit{attention weights} to produce a final representation. 

Let $f_{i1}, \cdots, f_{ik}$ be the $k$ morphological features of $w_i$, and denote by $\textbf{f}_{i1}, \cdots, \textbf{f}_{ik}$ their corresponding \textit{feature embeddings}. As in \textsection\ref{sec:parser}, $\textbf{h}_i$ and $\textbf{h}_j$ are the encodings of $w_i$ and $w_j$, respectively. The morphological representation $\textbf{m}_j$ of $w_j$ is:
\begin{align}
    \textbf{m}_j = [\textbf{f}_{j1}, \cdots, \textbf{f}_{jk}]^\top \textbf{k}
\end{align}
where $\textbf{k}$ is a vector of attention weights: 
\begin{align}
\label{attention-vec}
    \textbf{k} = \textrm{softmax}([\textbf{f}_{j1}, \cdots, \textbf{f}_{jk}]^\top \textbf{V} \textbf{h}_i )
\end{align}
The intuition is that dependent $w_i$ can choose which morphological features of $w_j$ are most important when deciding whether $w_j$ is its head. Note that this model is asymmetric: a word only attends to the morphological features of its (single) parent, and not its (many) children, which may have different functions.
\footnote{This is a simple and much less computationally demanding variant of the model of \newcite{dozat-qi-manning:2017:K17-3}, which uses different views for each head/dependent role.}

We combine the morphological representation with the word's encoding via a sigmoid gating mechanism.
\begin{align}
    \textbf{z}_j &= \textbf{g} \odot \textbf{h}_j + (1 - \textbf{g}) \odot \textbf{m}_j\\
    \textbf{g} & = \sigma(\textbf{W}_1 \textbf{h}_j + \textbf{W}_2 \textbf{m}_j)
\end{align}
where $\odot$ denotes element-wise multiplication. The gating mechanism allows the model to choose between the computed word representation and the weighted morphological representations, since for some dependencies, morphological features of the head might not be important. In the final model, we replace Eq. \ref{eq:head-pred} and Eq. \ref{eq:ass-score} with the following:
\begin{align}
    P_{head}(w_j|w_i, w) = \frac{\exp(a(\textbf{h}_i, \textbf{z}_j))}{\sum_{j'=0}^N \exp a(\textbf{h}_i, \textbf{z}_{j'})} \\
    a(\textbf{h}_i, \textbf{z}_j) = \textbf{v}_a \tanh(\textbf{U}_a \textbf{h}_i + \textbf{W}_a \textbf{z}_j)
\end{align}
The modified label prediction is:
\begin{align}
\label{eq:label-attn-pred}
P_{label}(\ell_k|w_i, w_j, w) = \frac{\exp(f(\textbf{h}_i, \textbf{z}_j)[k])}{\sum_{k'=0}^{|L|} \exp(f(\textbf{h}_i, \textbf{z}_{j})[k'])}
\end{align}
where $f$ is again a function to compute label score:
\begin{align}
\label{eq:lbl-attn-score}
f(\textbf{h}_i, \textbf{z}_j) = \textbf{V}_\ell \tanh(\textbf{U}_\ell \textbf{h}_i + \textbf{W}_\ell \textbf{z}_j)
\end{align}

\paragraph{Attention to headword morphological features}
We trained our augmented model (\textbf{oracle-attn}) on Finnish, German, Czech, and Russian. Its accuracy is very similar to the oracle model (Table \ref{tab:attn-res}), so we obtain a more interpretable model with no change to our main results.

\begin{table}[t]
\centering
\begin{tabular}{lcccc}
\toprule
\multirow{2}{*}{Language} & \multicolumn{2}{c}{oracle} & \multicolumn{2}{c}{oracle-attn} \\
 \cmidrule{2-5}
 & UAS & LAS & UAS & LAS \\
 \midrule
  Finnish & \textbf{89.2} & \textbf{87.3} & 88.9 & 86.9 \\
  Czech & 93.4 & 91.3 & \textbf{93.5} & \textbf{91.3} \\
  German  & 90.4 & 88.7 & \textbf{90.7} & \textbf{89.1} \\
  Russian & \textbf{93.9} & \textbf{92.8} & 93.8 & 92.7 \\
 \bottomrule
\end{tabular}
\caption{Our attention experiment results on development set.}
\label{tab:attn-res}
\end{table}

Next, we look at the learned attention vectors to understand which morphological features are important, focusing on the core arguments: nominal subjects, objects, and indirect objects. Since our model knows the case of each dependent, this enables us to understand what features it seeks in potential heads for each case. For simplicity, we only report results for words where both head and label predictions are correct.

\begin{figure}[t]
  \begin{subfigure}{\linewidth}
  \includegraphics[width=\linewidth]{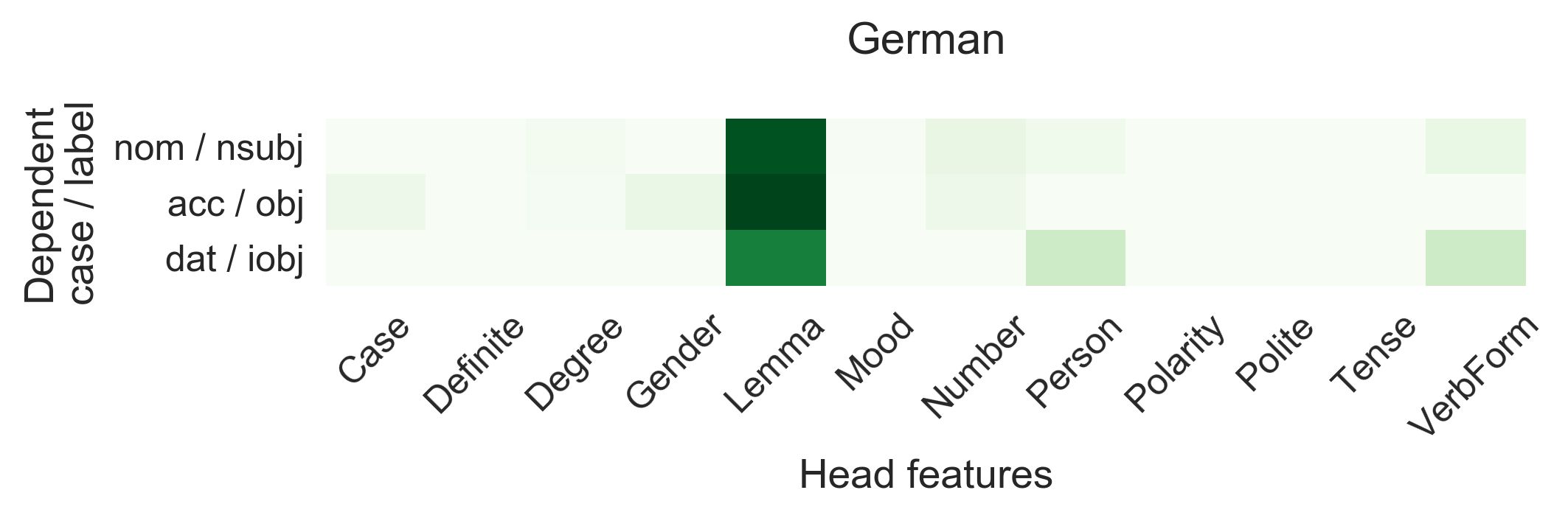}
  \end{subfigure}
  \begin{subfigure}{\linewidth}
  \includegraphics[width=\linewidth]{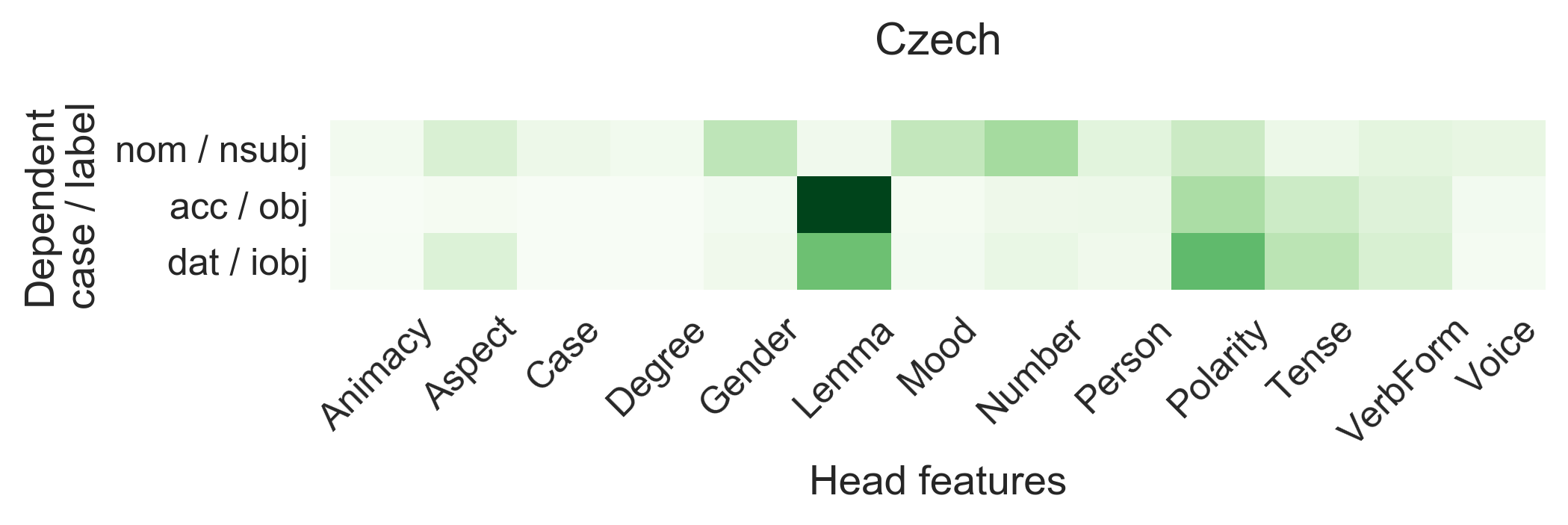}
  \end{subfigure}
  \begin{subfigure}{\linewidth}
  \includegraphics[width=\linewidth]{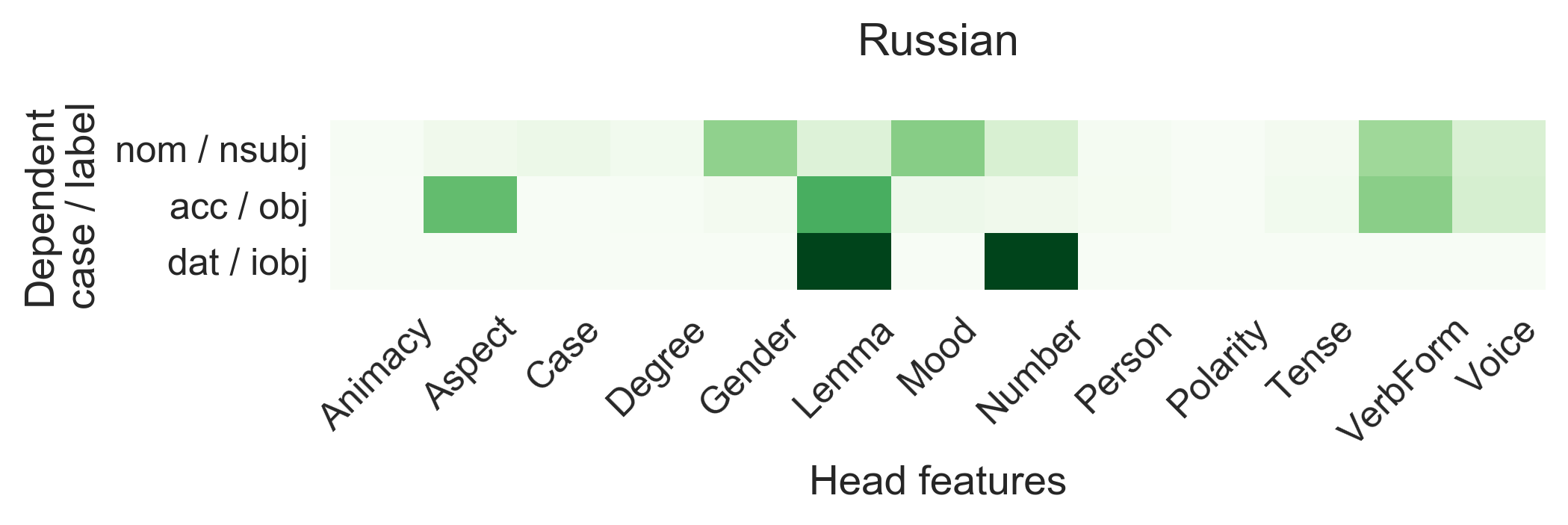}
  \end{subfigure}
\caption{The importance of morphological features of the head for subject and object predictions.}
\label{fig:attn-res}
\end{figure}

Figure~\ref{fig:attn-res} shows how attention is distributed across multiple features of the head word.
In Czech and Russian, we observe that the model attends to \textit{Gender} and \textit{Number} when the noun is in nominative case. This makes intuitive sense since these features often signal subject-verb agreement. As we saw in earlier experiments, these are features for which a character model can learn reliably good representations. For most other dependencies (and all dependencies in German),  \textit{Lemma} is the most important feature, suggesting a strong reliance on lexical semantics of nouns and verbs. However, we also notice that the model sometimes attends to features like 
\textit{Aspect}, \textit{Polarity}, and \textit{VerbForm}---since these features are present only on verbs, we suspect that the model may simply use them as convenient signals that a word is verb, and thus a likely head for a given noun.

\section{Conclusion}

Character-level models are effective because they can represent OOV words and orthographic regularities of words that are consistent with morphology. But they depend on context to disambiguate words, and for some words this context is insufficient. Case syncretism is a specific example that our analysis identified, but the main results in Table~\ref{tab:test-res} hint at the possibility that different phenomena are at play in different languages.

While our results show that prior knowledge of morphology is important, they also show that it can be used in a targeted way: our character-level models improved markedly when we augmented them only with case. This suggests a pragmatic reality in the middle of the wide spectrum between pure machine learning from raw text input and linguistically-intensive modeling: our new models don't need all prior linguistic knowledge, but they clearly benefit from some knowledge in addition to raw input. While we used a data-driven analysis to identify case syncretism as a problem for neural parsers, this result is consistent with previous linguistically-informed analyses \citep{seeker2013,Tsarfaty:2010:SPM:1868771.1868772}. We conclude that neural models can still benefit from linguistic analyses that target specific phenomena where annotation is likely to be useful.

\section*{Acknowledgments}

Clara Vania is supported by the Indonesian Endowment Fund for Education (LPDP), the Centre for Doctoral Training in Data Science, funded by the UK EPSRC (grant EP/L016427/1), and the University of Edinburgh. We would like to thank Yonatan Belinkov for the helpful discussion regarding morphological tagging experiments. We thank 
Sameer Bansal,
Marco Damonte,
Denis Emelin,
Federico Fancellu,
Sorcha Gilroy,
Jonathan Mallinson,
Joana Ribeiro,
Naomi Saphra, 
Ida Szubert,
Sabine Weber,
and the anonymous reviewers
for helpful discussion of this
work and comments on previous drafts
of the paper. 

\bibliography{emnlp2018}
\bibliographystyle{acl_natbib_nourl}

\clearpage
\appendix

\section{Supplemental Material}
\label{sec:supplemental}

\subsection{Morphological tagger}
\label{sec:morph-tagger}

We adapt the parser's encoder architecture for our morphological tagger. Following notation in Section \ref{sec:parser}, each word $w_i$ is represented by its context-sensitive encoding, $\textbf{h}_i$ (Eq. \ref{eq:enc-rep}). The encodings are then fed into a feed-forward neural network with two hidden layers---each has a ReLU non-linearity---and an output layer mapping the to the morphological tags, followed by a softmax. We set the size of the hidden layer to 100 and use dropout probability 0.2. We use Adam optimizer with initial learning rate 0.001 and clip gradients to 5. We train each model for 20 epochs with early stopping. The model is trained to minimized the cross-entropy loss.

Since we do not have additional data with the same annotations, we use the same UD dataset to train our tagger. To prevent overfitting, we only use the first 75\% of training data for training\footnote{We tried other settings, i.e. 25\%, 50\%, 100\%, but in general we achieve best result when we use 75\% of the training data.}. After training the taggers, we predict the case for the training, development, and test sets and use them for dependency parsing.

\subsection{Results on morphological tagging}
\label{sec:sup-tagging}

Table \ref{tab:classifier-res-de} and \ref{tab:classifier-res-ru} present morphological tagging results for German and Russian. We found that German and Russian have similar pattern to Czech (Table \ref{tab:classifier-res}), where morphological case seems to be preserved in the encoder because they are useful for dependency parsing. In these three fusional languages, contextual information helps character-level model to predict the correct case. However, its performance still behind the oracle. 

We observe a slightly different pattern on Finnish results (Table \ref{tab:classifier-res-fi}). The character embeddings achieves almost similar performance as the oracle embeddings. This results highlights the differences in morphological process between Finnish and the other fusional languages. We observe that performance of the encoder representations are slightly worse than the embeddings.

\begin{table}[H]
\centering
\setlength\tabcolsep{5pt}
  \begin{tabular}{lccccc}
  \toprule
  \multirow{2}{*}{Feature} & \multirow{2}{*}{baseline} & \multicolumn{2}{c}{embedding} & \multicolumn{2}{c}{encoder} 	\\
   \cmidrule{3-6}
   & & char & oracle & char & oracle \\
   \midrule
    Case  & 35.2 & 35.7 & \textbf{100} & 80.8 & \textbf{99.7} \\
    Gender & 56.8 & 63.6 & \textbf{100} & 75.7 & \textbf{78} \\
    Number & 59.1 & 67.1 & \textbf{100} & 78.3 & \textbf{93.9} \\
    All & 34 & 34.3 & \textbf{100} & 63.6 & \textbf{78.5} \\
   \bottomrule
  \end{tabular}
\caption{Morphological tagging results for German.}
\label{tab:classifier-res-de}
\end{table}

\begin{table}[H]
\centering
\setlength\tabcolsep{5pt}
  \begin{tabular}{lccccc}
  \toprule
  \multirow{2}{*}{Feature} & \multirow{2}{*}{baseline} & \multicolumn{2}{c}{embedding} & \multicolumn{2}{c}{encoder} 	\\
   \cmidrule{3-6}
   & & char & oracle & char & oracle \\
   \midrule
    Case  & 71.6 & 80.5 & \textbf{100} & 90.4 & \textbf{98.5} \\
    Gender & 87.7 & 97.4 & \textbf{100} & \textbf{69.9} & 57.3 \\
    Number & 83.7 & 94.5 & \textbf{100} & \textbf{85.7} & 83.8 \\
    All & 71.3 & 77.2 & \textbf{99.9} & \textbf{56.9} & 47.2 \\
   \bottomrule
  \end{tabular}
\caption{Morphological tagging results for Russian.}
\label{tab:classifier-res-ru}
\end{table}

\begin{table}[H]
\centering
\setlength\tabcolsep{5pt}
  \begin{tabular}{lccccc}
  \toprule
  \multirow{2}{*}{Feature} & \multirow{2}{*}{baseline} & \multicolumn{2}{c}{embedding} & \multicolumn{2}{c}{encoder} 	\\
   \cmidrule{3-6}
   & & char & oracle & char & oracle \\
   \midrule
    Case  & 56 & 96.7 & \textbf{100} & 88.9 & \textbf{91.4} \\
    Number & 56.4 & 97.4 & \textbf{100} & 81.9 & \textbf{89.5} \\
    All & 55.8 & \textbf{95} & 91.6 & 74 & \textbf{82.7} \\
   \bottomrule
  \end{tabular}
\caption{Morphological tagging results for Finnish.}
\label{tab:classifier-res-fi}
\end{table}

\end{document}